\documentclass[letterpaper, 10 pt, conference]{ieeeconf}
\IEEEoverridecommandlockouts
\usepackage{amsmath,amsfonts}
\usepackage{algorithm}
\usepackage{algpseudocode}
\usepackage{array}
\usepackage[caption=false,font=normalsize,labelfont=sf,textfont=sf]{subfig}
\usepackage{textcomp}
\usepackage{stfloats}
\usepackage{url}
\usepackage{verbatim}
\usepackage{graphicx}
\usepackage{xcolor}
\usepackage{multirow}

\usepackage{cite}
\begin{document}

\title{
From Biological Curves to Soft Modular Robots: Learning-Based Inverse Design with Uncertainty-Guided Refinement

}
\author{
Dixiang Wang\textsuperscript{*}, Minghao Zha\textsuperscript{*},
Arifur Rahman, Yinhong Qin, Yitao Jiang, \\
Shiyang Lu, Zebing Mao, Muhao Chen, Luyang Zhao%
\thanks{*Dixiang Wang and Minghao Zha contributed equally to this work.}%
\thanks{This work has been submitted to the IEEE for possible publication. Copyright may be transferred without notice, after which this version may no longer be accessible.}
}

\maketitle

\begin{abstract}
    Animals exhibit diverse shape-changing behaviors, from the undulatory swimming of eels to the arching movements of inchworms. Reproducing these biological shapes with modular soft robots offers a promising approach to developing versatile bioinspired robotic systems. However, translating a desired shape into a feasible robotic configuration remains challenging. Existing cable-driven modular soft robots rely on computationally expensive physics-based modeling and combinatorial search for inverse design, making it difficult to efficiently reproduce diverse target shapes. To address this challenge, we present a learning-based real-to-sim-to-real framework that translates biological curves into robotic configurations through surrogate-assisted inverse design and uncertainty-guided refinement. The proposed method combines learned inverse prediction with targeted candidate exploration, using a fast forward surrogate to evaluate and select suitable designs. The learned forward surrogate reduces deformation prediction time from 1.53~$\times$~10\textsuperscript{5} ms to 0.0008 ms, achieving approximately a 1.9~$\times$~10\textsuperscript{8}-fold speedup. At the same time, the proposed inverse design method reduces design time from 1.99~$\times$~10\textsuperscript{6} ms to 4.18 ms. On 5,000 test samples generated from previously unseen threading patterns, the method achieves a 93\% shape reconstruction success rate under surrogate-based evaluation. Physical demonstrations with four bioinspired robots further illustrate the translation of biological curves into realizable robotic shapes.
\end{abstract}

\section{Introduction}

\begin{figure*}[htbp]
    \centering
\includegraphics[width=\linewidth]{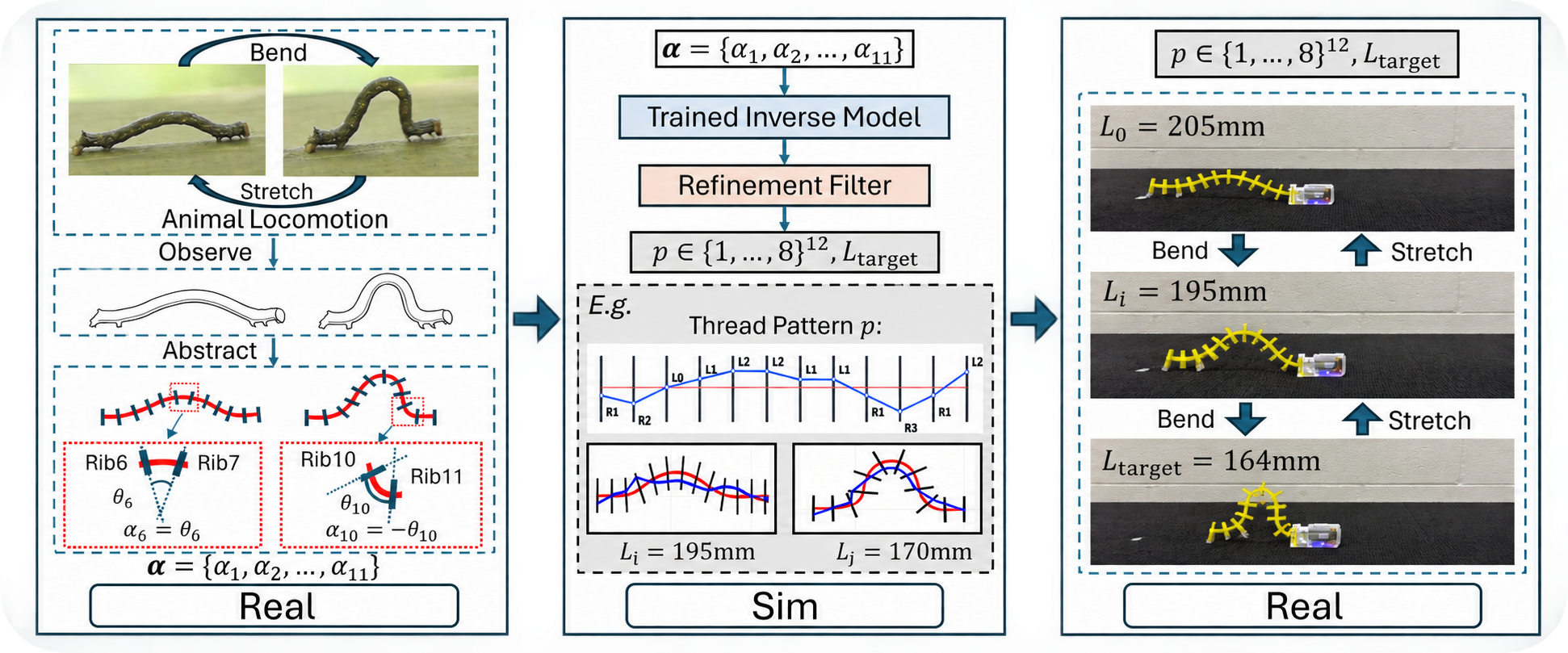}
    \caption{Overview of the animal-inspired real-to-sim-to-real design framework.
    Left: Key frames are extracted from real videos, and the curve is projected onto our skeleton to obtain an angle sequence. Middle: Our trained inverse model, combined with the refinement filter, maps the angle sequence to a threading pattern and target string length. Right: The threading pattern and target string length are applied to the physical skeleton, which is controlled to imitate the creature’s motion.}
    \label{Full_frame}
    \vspace{-1em}
\end{figure*}

Nature exhibits a rich diversity of shape-changing behaviors, from the swimming motions of eels~\cite{tytell2004eel} to the slithering of snakes~\cite{Jayne2020} and the arching movement of inchworms~\cite{11697326}.
Despite their distinct biological mechanisms, many of these deformations can be represented as geometric curves, providing a common description of diverse biological shapes~\cite{Ijspeert2014,hu2018small}. 
This observation motivates a real-to-sim-to-real approach to bio-inspired robot design, in which shapes observed in nature are abstracted into target curves, translated into feasible robotic configurations through simulation, and reproduced by physical robots, as illustrated in Fig.~\ref{Full_frame}. Soft robots are particularly suitable for this purpose, as their compliant structures enable flexible and adaptable deformations \cite{rus2015design,dong2022recent,Cecilia2016,zhao2025tensegrity}. However, translating a desired geometric curve into a physically realizable robotic shape requires identifying appropriate structural configurations and actuation parameters~\cite{10296178,dWebsterIII2010,9312179}.

SoftSnap, a previously developed cable-driven modular soft robotic platform, provides one possible approach to realizing such target curves \cite{zhao2025softsnap}. Each untethered module integrates onboard actuation, control, and power, enabling different bending curves through variations in cable-threading patterns and cable lengths. Multiple modules can be assembled using customized snap-together connectors to construct more complex robotic shapes. To predict the resulting deformation, SoftSnap employs a physics-based quasi-static forward model that uses iterative numerical optimization to minimize bending energy under a cable-length constraint. Although this model captures the deformation behavior, repeated forward evaluations can be computationally expensive {(152.75 s in average)}. The original platform also supports inverse design through a search-based approach that enumerates possible threading configurations, filters geometrically infeasible candidates, and evaluates the remaining configurations using the physics-based forward model. However, the number of possible threading configurations grows combinatorially with the number of ribs and available holes. Consequently, the inverse design process becomes computationally expensive due to both the large number of candidate configurations and the repeated numerical optimization required for forward evaluation. Restricting the number of evaluated candidates can reduce computational cost but may overlook configurations that better reproduce the target shape.

Learning-based methods offer a promising approach to reducing the computational cost of repeated physics-based evaluations. Previous studies have used neural networks to predict soft robot deformation \cite{10740917,9116011,10930705}, estimate cable tensions for desired tip positions in soft manipulators~\cite{giorelli2015inverse}, and perform inverse design using learned forward surrogates \cite{10816482,9726865,Yao2024,pmlr-v305-yi25a}. 
However, inverse design for the SoftSnap platform presents additional challenges due to its mixed discrete-continuous design space and non-unique inverse mapping \cite{Ansari2022}. Specifically, different combinations of cable-threading patterns and cable lengths can produce identical or nearly identical shapes, represented by their bending-angle vectors. This one-to-many relationship makes direct inverse prediction challenging~\cite{Liu2018,ardizzone2019analyzinginverseproblemsinvertible}. Conventional regression methods, such as those using mean squared error, may average cable lengths associated with different valid solutions, while selecting the most probable hole at each rib may combine threading choices from different feasible configurations. As a result, the predicted threading pattern and cable length may fail to reproduce the target shape accurately. Exhaustively evaluating alternative configurations, however, would reintroduce substantial computational costs. These challenges motivate an approach that combines fast inverse prediction with targeted candidate exploration and efficient surrogate-based evaluation.

To enable efficient inverse design within this real-to-sim-to-real workflow, we propose a surrogate-assisted approach that combines learning-based prediction with uncertainty-guided candidate refinement. First, a large-scale dataset is generated using the physics-based quasi-static model and curated to remove invalid and unstable samples. A forward surrogate is then trained to efficiently predict robot deformation given a cable-threading pattern and target cable length. Next, an inverse neural network is trained to jointly predict the cable-threading pattern and cable length from a target curve represented by a bending-angle vector. The inverse model is initially trained using supervised classification and regression losses, followed by end-to-end fine-tuning through a pretrained, frozen forward surrogate to minimize angular reconstruction error. Once trained, the inverse model generates an initial design, which is further improved through uncertainty-guided candidate refinement. Specifically, uncertain threading positions are identified using classification-logit margins, and alternative configurations are explored with adaptive search budgets. For more uncertain cases, local cable-length adjustments are also introduced. The forward surrogate then evaluates the resulting candidates and selects the configuration with the lowest predicted reconstruction error. Finally, physical demonstrations of biologically inspired soft robots illustrate the framework's potential for translating target geometric curves into realizable robotic shapes.

The main contributions of this work are summarized as follows: (a) A data-driven forward modeling framework that integrates physics-based data generation, data curation, and a learned forward surrogate to efficiently predict robot deformation without repeated numerical optimization during inference.
(b) A surrogate-assisted inverse design approach that combines learning-based inverse prediction with uncertainty-guided candidate refinement to selectively explore alternative cable-threading patterns and cable lengths, improving target-shape reconstruction without exhaustive search.

\section{Problem Formulation}

\begin{figure}[htbp]
    \centering
    \includegraphics[width=0.95\linewidth]{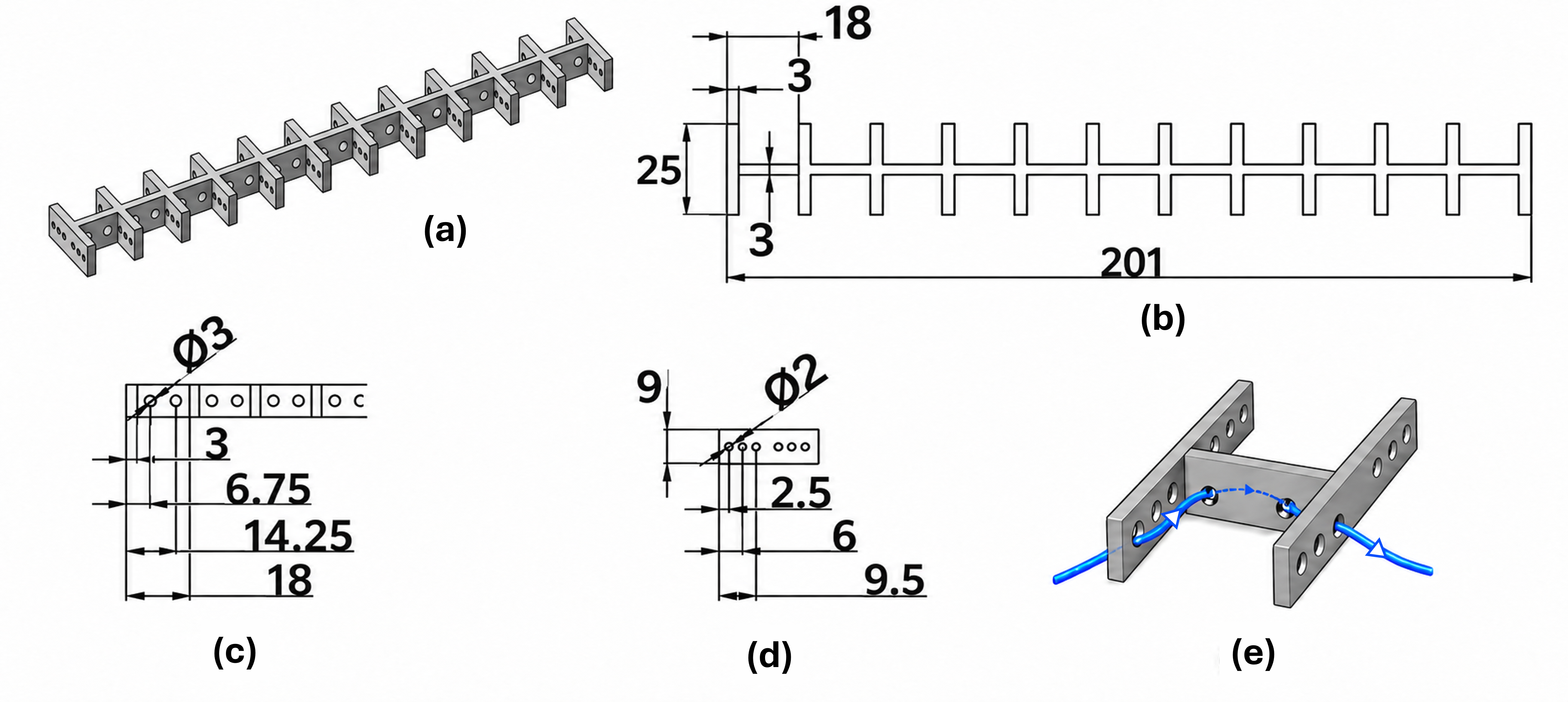}
\caption{Geometry and cable routing of a SoftSnap module (dimensions in mm). (a) Overview of the 12-rib skeleton. (b) Top view showing the overall dimensions and rib spacing. (c) Geometry and positions of the six threading holes on each rib. (d) Geometry of the two guide holes on the central spine between adjacent ribs. (e) Example of an R3-to-R0 cable-routing transition.}

    \label{Phy_Model}
\end{figure}

This study focuses on the quasi-static deformation of a single cable-driven soft robotic module from the previously developed SoftSnap platform \cite{zhao2025softsnap}. As illustrated in Fig.~\ref{Phy_Model}, the module consists of 12 ribs arranged along a flexible central spine and actuated by a single cable. By varying the cable-threading pattern and shortening the cable, relative rotations are induced between adjacent ribs, producing different bending shapes.

Each rib contains six physical threading holes, labeled L1–L3 and R1–R3, with two additional guide holes located along the central spine between adjacent ribs. The cable-threading pattern is denoted by $\mathbf{p}=(p_1,\ldots,p_{12})$, where $p_i\in\mathcal{P}=\{L0,L1,L2,L3,R0,R1,R2,R3\}$ specifies one of the eight routing choices at the $i$th rib. Unlike the six physical threading holes, L0 and R0 represent effective near-center routing choices designed to produce bending angles close to zero between adjacent ribs.
In the physical system, these near-center routing choices are implemented using the innermost rib holes and the two guide holes along the central spine. The cable passes through the first guide hole to the opposite side of the rib and then through the second guide hole to return to its original side. For example, an R3-to-R0 transition is implemented using the R1 hole and the two guide holes. Similarly, an R0-to-L0 transition is physically implemented as an R0-to-R0 routing, while an L0-to-R0 transition is implemented as an L0-to-L0 routing, preserving the original cable-routing side.

The target cable length is denoted by $L$. The resulting shape is represented by an 11-dimensional bending-angle vector $\boldsymbol{\alpha}=(\alpha_1,\ldots,\alpha_{11})$, where $\alpha_i$ denotes the relative bending angle between the $i$th and $(i+1)$th ribs.

\subsection{Physics-Based Forward Modeling}

The forward problem predicts the resulting robot shape given a cable-threading pattern $\mathbf{p}$ and target cable length $L$. This mapping is expressed as:
\begin{equation}
    \boldsymbol{\alpha}
    = f_{\mathrm{phys}}(\mathbf{p},L),
    \label{forward_model}
\end{equation}
where $f_{\mathrm{phys}}$ denotes the physics-based quasi-static forward model.
Following the original SoftSnap model \cite{zhao2025softsnap}, the equilibrium configuration is obtained by minimizing the bending energy subject to a fixed total cable-length constraint:
\begin{equation}
\begin{aligned}
    \min_{\boldsymbol{\alpha}} \quad &
    \sum_{i=1}^{11}\alpha_i^2 
    \quad\mathrm{s.t.}\quad &
    \sum_{i=1}^{11}
    \ell_i=L,
\end{aligned}
\label{physics_model}
\end{equation}
where $\ell_i$ denotes the cable-segment length between the $i$th and $(i+1)$th ribs, determined by their threading choices and relative bending angle. The resulting equilibrium bending-angle vector $\boldsymbol{\alpha}$ defines the output of the physics-based forward model.

\subsection{Inverse Design Problem}

Given a target shape represented by a bending-angle vector $\boldsymbol{\alpha}^*$, the inverse design problem seeks a cable-threading pattern $\mathbf{p}$ and target cable length $L$ that reproduce the desired deformation. Unlike the forward problem, the inverse mapping is \textbf{non-unique}: different combinations of threading patterns and cable lengths may produce identical or nearly identical bending-angle vectors. Moreover, some target shapes may be physically unattainable due to geometric and actuation constraints.
We therefore formulate inverse design as a constrained optimization problem:
\begin{equation}
(\mathbf{p}^*,L^*) \in
\arg\min_{(\mathbf{p},L)\in\mathcal{D}}
\mathrm{RMSE}\left(
f_{\mathrm{phys}}(\mathbf{p},L),
\boldsymbol{\alpha}^*
\right),
\label{inverse_model}
\end{equation}
where $\mathcal{D}$ denotes the admissible design space containing feasible threading configurations and cable lengths. The angular reconstruction error is defined as
$
\mathrm{RMSE}(\boldsymbol{\alpha},\boldsymbol{\alpha}^*)
=
\sqrt{\frac{1}{11}
\sum_{i=1}^{11}
(\alpha_i-\alpha_i^*)^2}$.

The objective is to identify a feasible design that accurately reproduces the target shape, rather than recovering a unique set of design parameters. However, directly solving this problem requires searching a large mixed discrete-continuous design space and repeatedly evaluating the physics-based forward model. This motivates the surrogate-assisted inverse design approach presented in the following section.

\section{Methodology}

The proposed framework consists of four main components: physics-based data generation and curation, forward surrogate modeling, two-stage inverse model training, and uncertainty-guided candidate refinement.

\subsection{Data Generation and Curation}

To train the forward and inverse models, a large-scale dataset is generated using the physics-based quasi-static model described in Section II. Each sample consists of a cable-threading pattern $\mathbf{p}$, a target cable length $L$, and the corresponding bending-angle vector $\boldsymbol{\alpha}$.
For each threading pattern, the cable is progressively shortened from its initial length $L_{\max}$ through a sequence of contraction steps. At each step, the constrained optimization problem in Eq.~\ref{physics_model} is solved using the equilibrium configuration from the previous step as the initial guess. This procedure generates multiple deformation states for each threading pattern.
However, the physics-based optimization does not explicitly enforce all geometric feasibility constraints. Consequently, some generated states may exhibit excessive bending, collisions between adjacent ribs, or numerical instability. To improve dataset quality, a curation process is introduced to remove geometrically invalid, redundant, and unstable samples.

\begin{figure}[htbp]
    \centering
    \includegraphics[width=\linewidth]{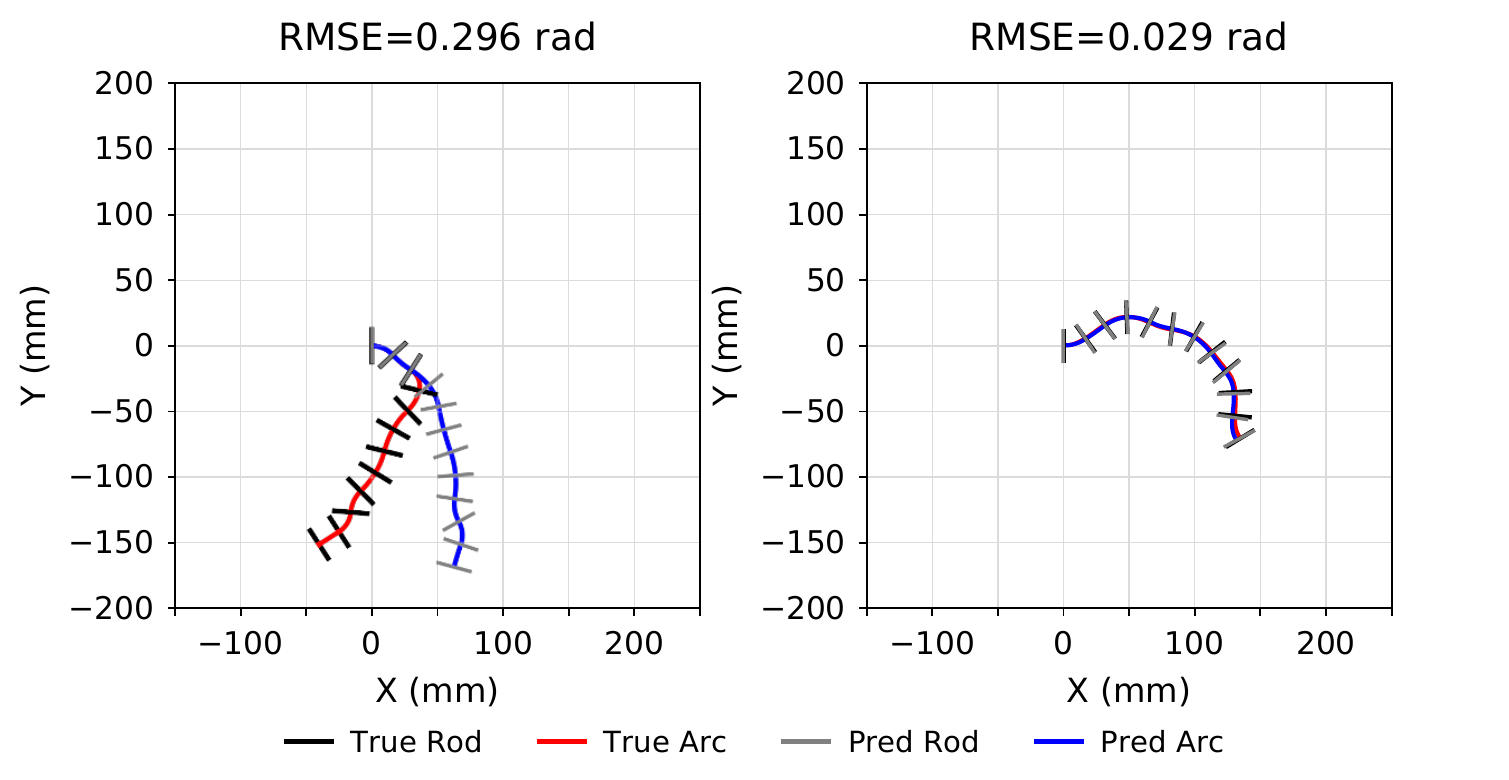}
    \caption{Illustrations of $0.2 <$ RMSE and RMSE $< 0.03$. Black and gray line segments represent ground-truth and predicted rods, respectively; red and blue curves represent the corresponding ground-truth and predicted arcs.}
    \label{0.03-0.2}
\end{figure}

First, each bending angle is constrained to $\alpha_i \in [-1.5,1.5]~\mathrm{rad}$ to exclude configurations with excessive bending that may cause collisions between adjacent ribs. In addition, changes in bending-angle vectors between consecutive contraction steps are measured as
$
d_t =
\sqrt{
\frac{1}{11}
\sum_{i=1}^{11}
\left(
\alpha_i^{(t)}-\alpha_i^{(t-1)}
\right)^2
}.
\label{data_filter}
$ As shown in Fig. \ref{0.03-0.2}, samples with $d_t<0.03$ rad are considered redundant due to negligible deformation changes, while samples with $d_t\geq0.2$ rad are treated as potentially numerically unstable. Only samples satisfying $0.03\leq d_t<0.2$ rad are retained; otherwise, the remaining contraction steps for the current threading pattern are terminated.

\subsection{Forward Surrogate Modeling}

\begin{figure*}[htpb]
    \centering
    \includegraphics[width=\linewidth]{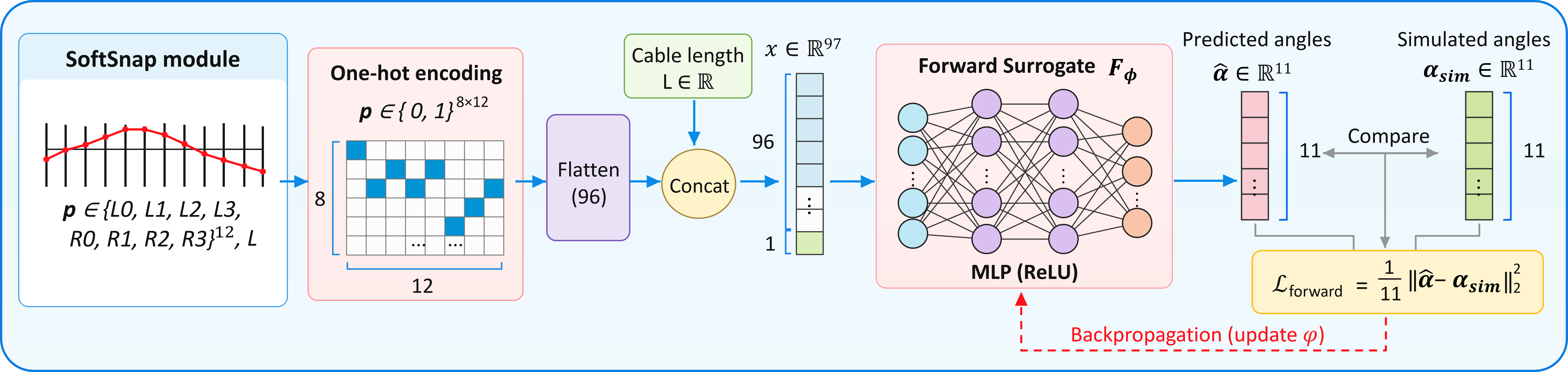}
    \caption{Forward surrogate training framework.}
    \label{fig:Forwar_training}
\end{figure*}

The forward surrogate approximates the physics-based mapping from cable-threading patterns and cable lengths to bending-angle vectors, avoiding repeated numerical optimization during inference.
As illustrated in Fig.~\ref{fig:Forwar_training}, the discrete threading pattern is represented using one-hot encoding. Each of the 12 rib positions has eight possible routing choices, corresponding to L0--L3 and R0--R3 as defined in Section II. The pattern is encoded as a matrix $\mathbf{P}\in\{0,1\}^{8\times12}$,  where each row represents the selected routing choice at one rib.
The matrix is flattened and concatenated with the continuous cable length $L$ to form the forward model input. 
Before concatenation, $L$ is standardized using z-score normalization as follows:
\begin{equation}
    L_{input}=\frac{L-\mu_{L}}{\sigma_L+10^{-8}}
\end{equation}
where $\mu_L$ and $\sigma_L$ denote the mean and standard deviation, respectively, of the string lengths in the training set.


A multilayer perceptron (MLP) is employed to learn the nonlinear mapping between the design parameters and the resulting deformation:
\begin{equation}
\hat{\boldsymbol{\alpha}}
=F_{\phi}(\mathbf{P},L),
\label{surrogate_model}
\end{equation}
where $\phi$ denotes the trainable parameters of the forward surrogate and $\hat{\boldsymbol{\alpha}}\in\mathbb{R}^{11}$ is the predicted bending-angle vector.

The forward model is trained using the simulated bending-angle vectors as supervision. The training loss is defined as the mean squared error between the predicted and simulated bending angles:
\begin{equation}
{
\mathcal{L}_{\mathrm{forward}}
=
\frac{1}{11}
\left\|
F_{\phi}(\mathbf{P},L)
-\boldsymbol{\alpha}
\right\|_2^2.
}
\label{forward_loss}
\end{equation}

Once trained, the forward surrogate provides efficient deformation predictions and is used for both inverse model fine-tuning and candidate evaluation.

\subsection{Two-Stage Inverse Model Training}

Given a target bending-angle vector $\boldsymbol{\alpha}^*$, the inverse model aims to jointly predict the corresponding cable-threading pattern and target cable length. Since the design parameters include both discrete and continuous variables, the inverse model performs classification and regression.

The inverse network is expressed as:
\begin{equation}
(\mathbf{Z},\hat{L})
=
I_{\theta}(\boldsymbol{\alpha}^*),
\label{inverse_network}
\end{equation}
where $\theta$ denotes the trainable parameters of the inverse network, $\mathbf{Z}\in\mathbb{R}^{12\times8}$ contains the routing classification logits, and $\hat L$ is the predicted cable length.

The inverse model is trained in two stages.

\textit{Stage 1: Supervised Pretraining.}
The inverse network is first trained using the threading patterns and cable lengths from the generated dataset. Cross-entropy loss is employed for routing classification, while a regression loss is used for cable-length prediction. The supervised training objective is expressed as:
\begin{equation}
\mathcal{L}_{\mathrm{sup}}
=
\frac{1}{12}
\sum_{i=1}^{12}
\mathrm{CE}(\mathbf{Z}_i,p_i)
+
\lambda_L
\mathcal{L}_{\mathrm{length}}(\hat L,L),
\label{inverse_supervised_loss}
\end{equation}
where $\mathrm{CE}$ denotes the cross-entropy loss for the categorical routing choices, and $\lambda_L=1$ balances the classification and regression objectives.

\textit{Stage 2: End-to-End Fine-Tuning.}
To directly optimize shape reconstruction accuracy, the pretrained forward surrogate is incorporated into the inverse training process. During this stage, the forward surrogate parameters $\phi$ are frozen, while the inverse model parameters $\theta$ remain trainable.

To maintain differentiability, the predicted routing logits are converted into a continuous probability matrix using Softmax:
\begin{equation}
P_{i,k}^{\mathrm{soft}}
=
\frac{\exp(Z_{i,k})}
{\sum_{j=1}^{8}\exp(Z_{i,j})}.
\label{inverse_softmax}
\end{equation}

The resulting probability matrix and predicted cable length are passed through the frozen forward surrogate:
\begin{equation}
\hat{\boldsymbol{\alpha}}
=
F_{\phi}
(\mathbf{P}^{\mathrm{soft}},\hat L).
\label{inverse_reconstruction}
\end{equation}

The angular reconstruction loss is defined as:
\begin{equation}
\mathcal{L}_{\mathrm{recon}}
=
\frac{1}{11}
\left\|
\hat{\boldsymbol{\alpha}}
-
\boldsymbol{\alpha}^*
\right\|_2^2.
\label{inverse_reconstruction_loss}
\end{equation}

The reconstruction loss is backpropagated through the frozen forward surrogate to update the inverse network. This allows the inverse model to be optimized according to the resulting deformation rather than solely matching the design parameters recorded in the training dataset.


\begin{figure*}[htbp]
    \centering
    \includegraphics[width=\linewidth]{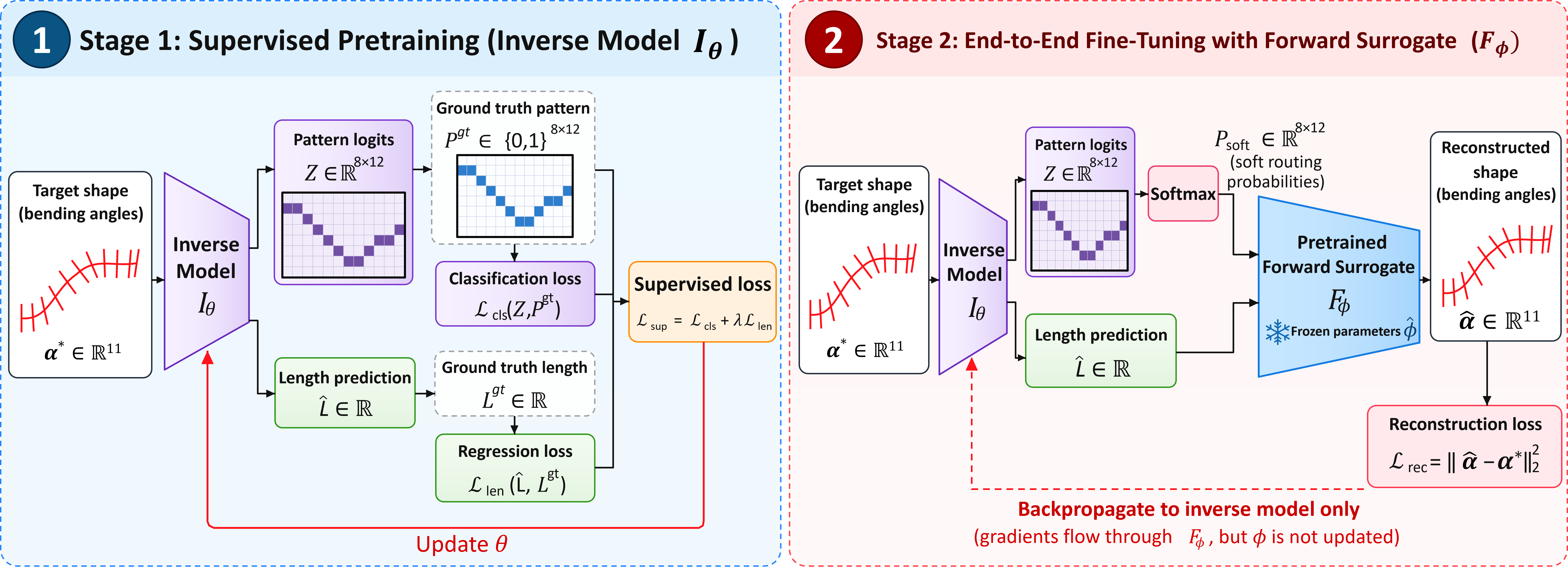}
\caption{Two-stage inverse model training. Stage 1 performs supervised pretraining using a combined classification and regression loss. Stage 2 minimizes angular reconstruction error through a pretrained, frozen forward surrogate, updating only the inverse model parameters.}
    \label{Inverse_flowchart}
    \vspace{-1em}
\end{figure*}

\subsection{Uncertainty-Guided Candidate Refinement}

Once trained, the inverse model generates an initial threading pattern and cable length for a given target shape. However, due to the non-unique inverse mapping, directly accepting a single predicted configuration may result in inaccurate shape reconstruction.

To improve reconstruction accuracy without exhaustive search, an uncertainty-guided candidate refinement strategy is introduced. Rather than exploring the entire design space, this strategy focuses on routing positions where the inverse model exhibits greater prediction ambiguity.


For each rib position $i$, the inverse network produces eight classification logits $Z_{i,k}$. The uncertainty of the routing prediction is estimated using the gap between the two largest logits:
\begin{equation}
g_i=Z_{i,(1)}-Z_{i,(2)},
\label{logit_gap}
\end{equation}
where $Z_{i,(1)}$ and $Z_{i,(2)}$ denote the largest and second-largest logits, respectively. A smaller gap indicates greater ambiguity between the two most likely routing choices.
We select the $R$ ribs with the highest uncertainty scores $g_i$ for further search. For each selected rib, we then choose the top-$k$ candidate holes based on $Z_{i,(k)}$. The resulting search space contains $k^R$ candidate configurations. 

In addition to threading-pattern refinement, local cable-length adjustments are introduced for hard samples. The candidate lengths are generated as:
\begin{equation}
L^{(j)}=\hat L+\Delta L,
\quad
\Delta L\in\{0,\pm4~mm,\pm8~mm\}.
\label{length_perturbation}
\end{equation}

The resulting threading patterns and cable lengths form a candidate set $\mathcal{C}$ that includes the initial inverse prediction.

For each candidate $(\mathbf{p}^{(j)},L^{(j)})\in\mathcal{C}$, the learned forward surrogate reconstructs the corresponding bending-angle vector:
\begin{equation}
\hat{\boldsymbol{\alpha}}^{(j)}
=
F_{\phi}
\left(
\mathrm{onehot}(\mathbf{p}^{(j)}),
L^{(j)}
\right).
\label{candidate_forward}
\end{equation}

The final configuration is selected as the candidate with the lowest surrogate-predicted angular reconstruction error:
\begin{equation}
(\mathbf{p}^{\mathrm{ref}},L^{\mathrm{ref}})
\in
\arg\min_{(\mathbf{p},L)\in\mathcal{C}}
\mathrm{RMSE}
\left(
F_{\phi}(\mathrm{onehot}(\mathbf{p}),L),
\boldsymbol{\alpha}^*
\right).
\label{candidate_selection}
\end{equation}

By selectively exploring uncertain routing positions and evaluating candidates through the learned forward surrogate, the proposed method refines the initial inverse prediction without exhaustively evaluating the combinatorial design space.

\section{Experiments}

The experiments evaluate three questions: (1) whether a learned forward surrogate can accurately and efficiently replace repeated physics-based deformation predictions; (2) whether the proposed inverse design method can reconstruct target shapes for previously unseen threading patterns; and (3) how supervised pretraining and uncertainty-guided candidate refinement contribute to inverse design performance. 

\subsection{Experimental Setup and Dataset}
\label{Dataset}
The dataset is generated using the physics-based quasi-static SoftSnap model \cite{zhao2025softsnap} and the data curation procedure described in Section III. Each sample contains a threading pattern $\mathbf{p}$, target cable length $L$, and corresponding bending-angle vector $\boldsymbol{\alpha}\in\mathbb{R}^{11}$. The training dataset contains approximately 5M samples from around 115k distinct threading patterns, where the data were split into training and validation sets at a ratio of 95:5. An independent test set of 5k samples is generated from threading patterns excluded from training. Since each pattern can be evaluated at multiple cable lengths, the pattern-level separation prevents exact threading configurations from appearing in both the training and test sets.

To examine data efficiency, models are additionally trained using subsets of 10k, 100k, 500k, 1M, and 3M samples, in addition to the full 5-million-sample dataset. The test set is held fixed across training sizes. 

The primary evaluation metric is the angular root-mean-square error (RMSE) defined in Section II. Shape reconstruction is considered successful when the error is below \(\epsilon = 0.03\) rad, as the reconstructed curves closely resemble the target curves within this threshold. For the forward task, predictions are compared with the equilibrium angles generated by the physics-based simulator. For the inverse task, the reported model-comparison results use the learned forward surrogate to reconstruct the angles produced by predicted threading patterns and cable lengths; these results are therefore referred to as \emph{surrogate-based inverse evaluations}.

Runtime is measured separately for a single forward prediction and for the complete inverse design procedure, including candidate generation and evaluation. All measurements are performed on a system equipped with an NVIDIA H200 GPU and an Intel Xeon Platinum 8580 CPU, using Python 3.11.16, PyTorch 2.11.0, and CUDA 12.8, with a batch size of 131,072. 

\subsection{Forward Surrogate Evaluation}
\label{sec:forward_results}

The forward task predicts the 11-dimensional bending-angle vector from a threading pattern and cable length. Six model families are compared: $k$-nearest neighbors (kNN), CNN, XGBoost, Transformer encoder, Transformer decoder, and MLP. All models are evaluated on the held-out test set. Table~\ref{tbl:model_comparison} summarizes their accuracy, reported inference time, and model size. Among the evaluated learning-based models, the MLP obtains the lowest angular RMSE (0.0077 rad) and highest success rate (96.02\%), and is therefore selected as the forward surrogate used in subsequent inverse design experiments. Compared with the Transformer decoder, its RMSE is reduced from 0.0186 to 0.0077 rad. Compared with the original physics-based forward optimizer, which requires 152.75 seconds for a matched $(\mathbf{p},L)$ query, the learned MLP surrogate requires 0.0008 milliseconds, corresponding to a $1.9\times10^8$-fold speedup. Increasing the training set from 10k to 5M samples improved prediction success from approximately 74\% to 93\%, with performance gradually saturating at approximately 96\% as the dataset size increased to 5 million samples.

\begin{table}[!htbp]
\caption{Performance Comparison of Forward Surrogate Models.}
\centering
\resizebox{\columnwidth}{!}{%
\begin{tabular}{c|cccccc}
\hline
\textbf{Model} & \textbf{RMSE (rad)} & \textbf{Success Rate} & \textbf{Time (ms)} & \textbf{Size (MB)} \\
\hline
kNN (Baseline) & 0.2016 & 0.0778 & 12.2777 & 231.9694 \\
CNN & 0.0219 & 0.8454 & 0.1534 & 1.194 \\
XGBoost & 0.0346 & 0.6520 & 47.6386 & 206.38 \\
Transformer (Enc) & 0.0325 & 0.7126 & 0.0490 & 0.067 \\
Transformer (Dec) & 0.0186 & 0.8692 & 0.0471 & 0.148 \\
MLP (Proposed) & \textbf{0.0077} & \textbf{0.9602} & \textbf{0.0008} & 130.592 \\
\hline
\end{tabular}%
}
\label{tbl:model_comparison}
\end{table}


\subsection{Inverse Design Evaluation}
\label{sec:inverse_results}

Inverse design predicts a threading pattern and cable length from a target bending-angle vector. Since multiple parameter configurations can reproduce similar shapes, the evaluation focuses on the reconstructed shape instead of exact recovery of the original threading labels. The returned discrete configurations are evaluated using the forward surrogate and the angular RMSE defined in Section II.

\begin{figure}[htbp]
    \centering
    \vspace{-1em}
    \includegraphics[width=\linewidth]{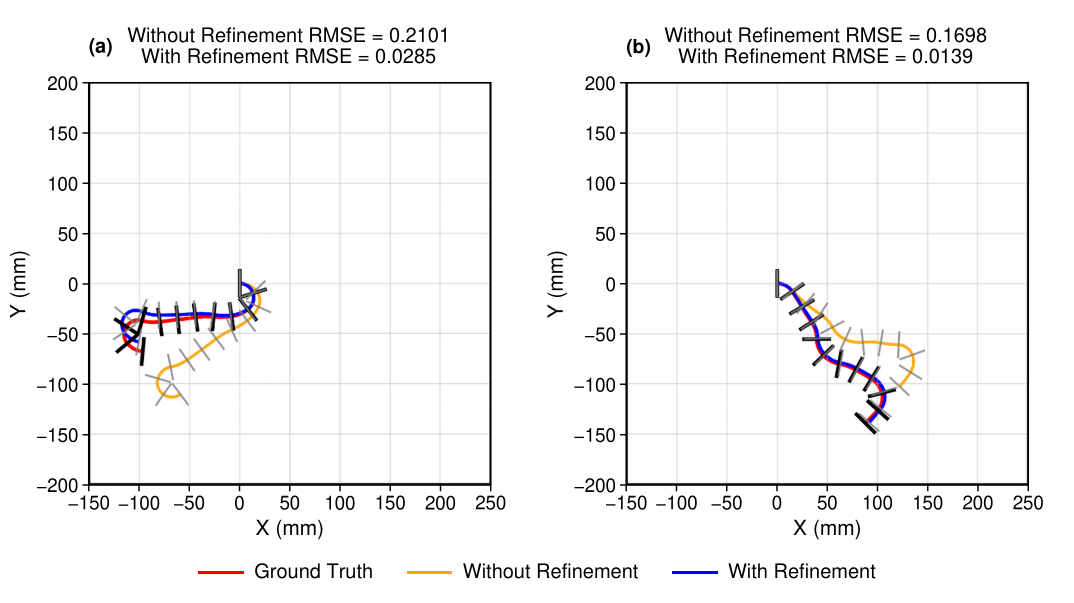}
    \caption{Two test examples with the largest improvement in shape reconstruction error from initial prediction to uncertainty-guided candidate refinement.}
    \label{Refinement}
    \vspace{-1em}
\end{figure}

Table~\ref{tbl:inverse_comparison} compares the original SoftSnap inverse design method with several learning-based approaches, including kNN, a decoder-only Transformer, and MLPs with different training strategies and candidate refinement.
The original SoftSnap method achieves an average runtime of 1.99$\times10^6$ ms per target and a reconstruction success rate of 100\%.
Among the supervised-only baselines, the decoder-only Transformer achieves the highest success rate of 78.08\%, which increases to 85.06\% with candidate refinement. The two-stage MLP achieves reconstruction success rates of 79.44\% without refinement and 93.18\% with candidate refinement. With $R=3$ and $k=6$, the refined model achieves the highest success rate among the methods compared in this table, with an average runtime of 4.182 ms per sample. 
Our experiments show that increasing $R$ and $k$ can further improve the success rate. For example, setting $R=6$ and $k=8$ increases the success rate to 94.48\%, but at a substantially higher runtime of 3967.094 ms per sample.
These results suggest that supervised initialization and candidate refinement provide complementary improvements in inverse design performance. Fig.~\ref{Refinement} illustrates two examples in which candidate refinement corrects inaccurate initial threading choices and improves shape reconstruction.

\begin{table*}[htbp]
\caption{Performance comparison of inverse design methods.}
\centering
\resizebox{\textwidth}{!}{%
\begin{tabular}{llcccc}
\hline
\textbf{Method} & \textbf{Search strategy} & \textbf{RMSE (rad)} & \textbf{Success} & \textbf{Time (ms)} & \textbf{Size (MB)} \\
\hline
Original SoftSnap inverse search & Physics-based search & 0.0973 & 1.000 & 1.99~$\times$~10\textsuperscript{6} & --- \\
kNN & None & 0.1297 & 0.1246 & 52.723 & 2059.937 \\
CNN & None & 0.0459 & 0.4496 & 0.220 & 0.532 \\
XGBoost & None & 0.0748 & 0.2438 & 0.212 & 59.354 \\
Transformer (Encoder)& None & 0.0413 & 0.5052 & 0.002 & 1.571 \\
Transformer (Decoder)& None & 0.0250 & 0.7808 & 0.084 & 4.871 \\
MLP (one-stage) & None & 0.0346 & 0.6290 & 0.016 & 132.426 \\
MLP (end-to-end) & None & 0.0304 & 0.6990 & 0.020 & 132.426 \\
MLP (two-stage) & None & 0.0230 & 0.7944 & 0.124 & 132.426 \\
Transformer (Decoder) + refinement& Uncertainty-guided & 0.0179 & 0.8506 & 3.602 & 4.871 \\
MLP (end-to-end) + refinement & Uncertainty-guided & 0.0150 & 0.8760 & 3.277 & 132.426 \\
\textbf{MLP (two-stage) + refinement [proposed]}& Uncertainty-guided & \textbf{0.0099} & \textbf{0.9318} & 4.182 & 132.426 \\
\hline
\end{tabular}%
}
\label{tbl:inverse_comparison}
\end{table*}

\section{Physical Validation}

\begin{figure*}[!t]
    \centering
    \includegraphics[width=\linewidth]{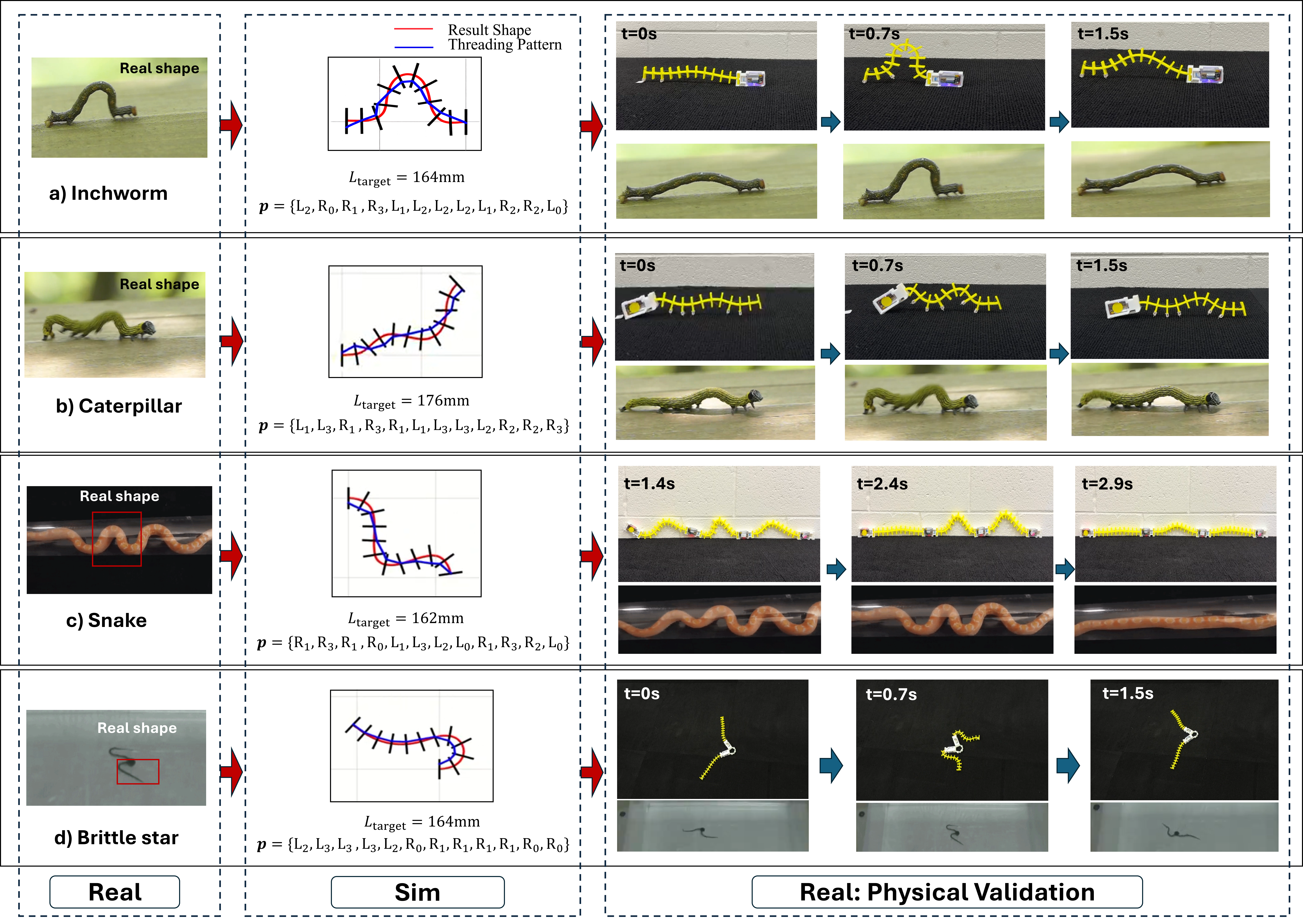}
    \caption{Real-to-sim-to-real physical validation of four bioinspired robots: (a) inchworm, (b) caterpillar, (c) snake, and (d) brittle star. Biological shapes are abstracted into target curves (Real), mapped to cable-threading patterns and target lengths through inverse design (Sim), and realized using physical SoftSnap modules (Real).}
    \vspace{-2em}
    \label{Inchworm_Ca}
\end{figure*}

To evaluate the physical realization of the proposed real-to-sim-to-real workflow, four bioinspired robot configurations are considered: inchworm, caterpillar, snake, and brittle star, as shown in Fig. \ref{Inchworm_Ca}. For each case, a representative frame of the biological motion is used to extract and normalize a target curve, which is represented by an 11-dimensional bending-angle vector. The trained inverse model, followed by uncertainty-guided candidate refinement, determines a cable-threading pattern and target cable length for reproducing the desired shape. The selected configuration is then implemented using SoftSnap modules. 

\begin{figure}[htbp]
    \centering
    \captionsetup[subfloat]{font=scriptsize,labelfont=scriptsize}
    \subfloat[]{
        \includegraphics[width=0.48\linewidth]{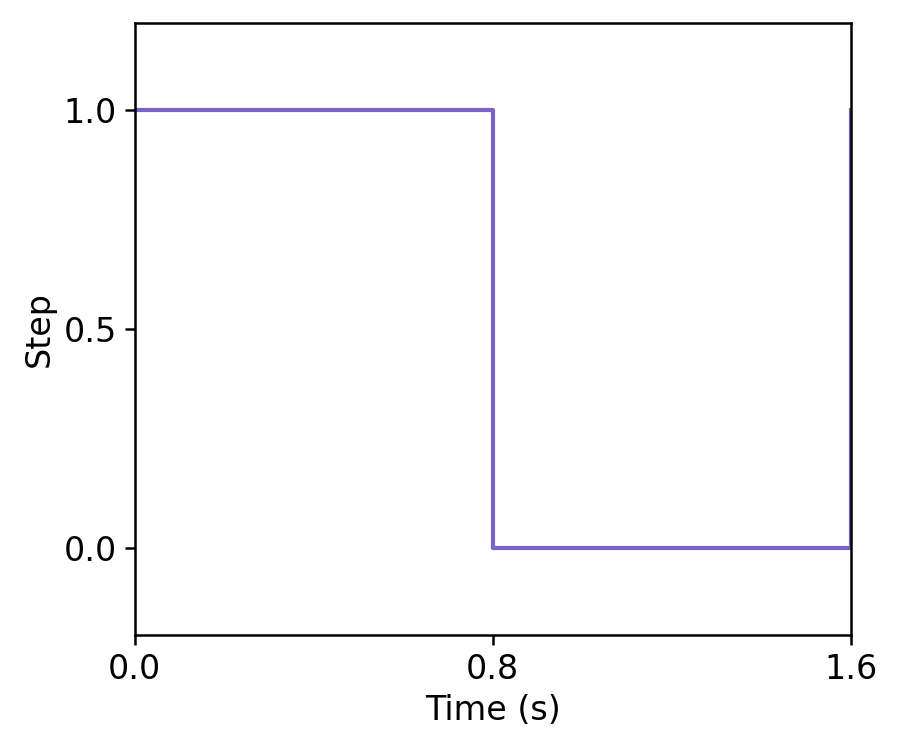}
        \label{Control_signal}
    }
    \subfloat[]{
        \includegraphics[width=0.48\linewidth]{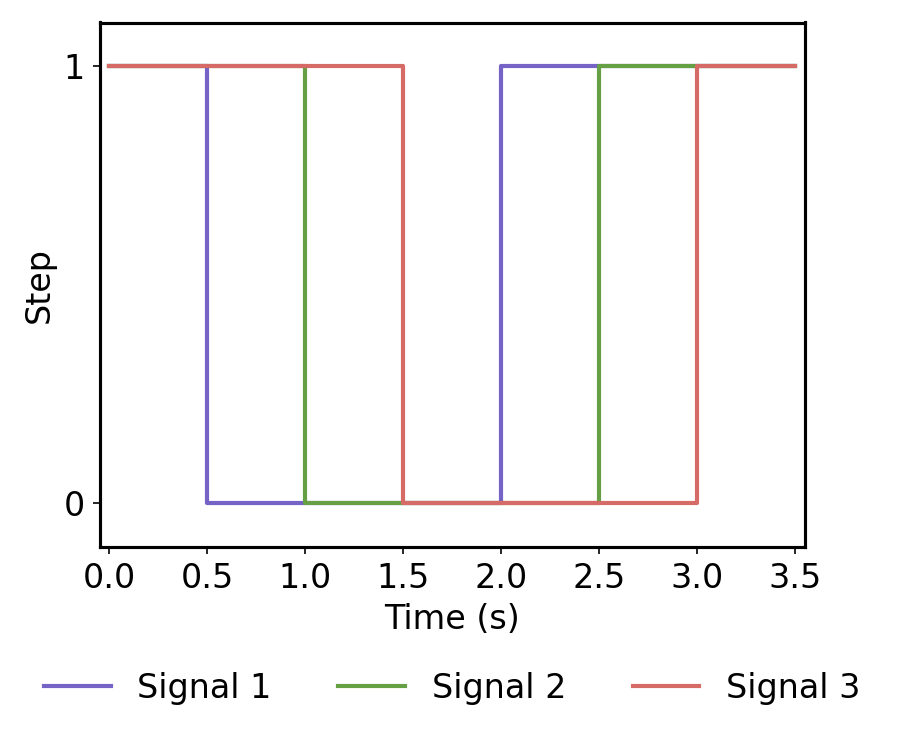}
        \label{Control_snake}
    }
    \caption{Motor command signals for physical validation. (a) Commands for the inchworm-, caterpillar-, and brittle star-inspired robots; the two brittle star motors receive the same signal. (b) Three independently commanded motors for the snake-inspired robot.}
    \label{fig:physical_control_signals}
    \vspace{-2em}
\end{figure}

To examine whether the resulting robotic configurations can generate locomotion, the physical robots are driven using prescribed cyclic motor commands. The inchworm- and caterpillar-inspired robots each use one motor, while the snake-inspired robot uses three modules with three independently commanded motors. The brittle star-inspired robot uses two motors driven by the same signal. The corresponding motor commands are shown in Fig.~\ref{fig:physical_control_signals}. These commands are specified for the locomotion demonstrations; the inverse design model predicts cable-threading patterns and target cable lengths rather than full time-varying control sequences.

Five locomotion trials were conducted for each robot, with three actuation cycles per trial for the inchworm and snake and five for the caterpillar and brittle star. Each cycle lasted 1.6 s, except for the snake (3.0 s). The average speeds were 4.25, 4.08, 1.09, and 9.50 cm/s, corresponding to 6.80, 6.52, 3.27, and 15.20 cm/cycle for the inchworm, caterpillar, snake, and brittle star, respectively. These results demonstrate locomotion capability across different robotic configurations.
These demonstrations illustrate the feasibility of translating biological target curves into physically realizable SoftSnap configurations capable of producing bioinspired deformations and locomotion. The evaluation focuses on representative shape realization and physical motion rather than on reproducing complete biological gait trajectories.



\section{Conclusion and Future Work}

This paper presented a learning-based real-to-sim-to-real framework for translating biological curves into physically realizable soft modular robots. Building on the SoftSnap platform, the proposed approach combines a learned forward surrogate with an inverse model and uncertainty-guided candidate refinement to address the computational challenges of physics-based optimization and combinatorial inverse search. This enables efficient inverse design without exhaustively searching all possible threading configurations.

Experiments on previously unseen threading patterns demonstrate that the proposed refinement improves the reconstruction success rate from 79.44\% to 93.18\% under surrogate-based evaluation. With an average runtime of 4.182 ms per sample, the proposed inverse design method achieves an approximately $4.76\times10^5$-fold speedup over the original physics-based approach. The learned forward surrogate further accelerates deformation prediction by more than eight orders of magnitude, enabling rapid candidate evaluation. Physical demonstrations with four bioinspired robots illustrate the feasibility of translating biological target curves into physically realizable soft modular robots. The current framework focuses on quasi-static shape reconstruction within the studied SoftSnap configuration space, without accounting for the effects of external forces such as gravity and friction or the robot's dynamics. Future work will extend the framework to incorporate these physical effects into dynamic simulations, enabling more realistic predictions of robot motion and interactions with the environment.

\bibliographystyle{IEEEtran}
\bibliography{references}










\newpage

\vfill

\end{document}